%% file: main.tex
\documentclass[11pt,a4paper]{article}
\usepackage[hyperref]{acl2020}
\usepackage{times}
\usepackage{latexsym}

\usepackage{microtype}

\usepackage{multirow}
\usepackage{amsmath}
\usepackage{booktabs}
\usepackage{url}
\usepackage{graphicx}

\usepackage{subcaption}

\usepackage{xcolor}
\usepackage{xparse}

\definecolor{color_s}{RGB}{32,128,32} 
\definecolor{color_f}{RGB}{144,32,144} 
\definecolor{color_c}{RGB}{96,96,96} 

\ExplSyntaxOn

\NewDocumentCommand{\fs}{m}
 {
  \tl_map_function:nN { #1 } \firth_fs:n
 }

\cs_new_protected:Nn \firth_fs:n
 {
  \str_case:nn { #1 }
   {
    {f}{\__firth_fs_box:nn { color_f } { f }}
    {s}{\__firth_fs_box:nn { color_s } { s }}
    {c}{\__firth_fs_box:nn { color_c } { c }}
    
   }
 }
\cs_new_protected:Nn \__firth_fs_box:nn
 { 
  {\setlength{\fboxsep}{0.5pt}\colorbox{#1}
   {
    \color{white}
    \use:c { check@mathfonts }
    \fontsize { \use:c{ sf@size } } { 0 } \selectfont
     \texttt{\textbf{#2}} \vphantom{f|} 
   }}
 }

\ExplSyntaxOff

\aclfinalcopy 

\title{Improving Transformer Models by Reordering their Sublayers}
\author{Ofir Press$^\diamondsuit$  \quad Noah A. Smith$^\diamondsuit$$^\spadesuit$ \quad Omer Levy$^\clubsuit$\\
\\ $^\diamondsuit$Paul G. Allen School of Computer Science \& Engineering, University of Washington
\\ $^\spadesuit$Allen Institute for AI
\\ $^\clubsuit$Facebook AI Research
}
\date{}

\begin{document}
\maketitle

\input{00-abstract.tex}
\input{01-intro.tex}
\input{02-background.tex}

\input{03-hypothesis.tex}

\input{04-design.tex}

\input{05-beyond.tex}
\input{06-analysis.tex}
\input{07-related.tex}
\input{08-conclusion.tex}

\section*{Acknowledgments}
We thank Tim Dettmers, Jungo Kasai,  Sainbayar Sukhbaatar, and the anonymous reviewers for their valuable feedback.

\bibliography{acl2020}
\bibliographystyle{acl_natbib}

\end{document}

%% file: 00-abstract.tex
\begin{abstract}
Multilayer transformer networks consist of interleaved self-attention and feedforward sublayers.
Could ordering the sublayers in a different pattern lead to better performance?
We generate randomly ordered transformers and train them with the language modeling objective.
We observe that some of these models are able to achieve better performance than the interleaved baseline, and that those successful variants tend to have more self-attention at the bottom and more feedforward sublayers at the top.
We propose a new transformer pattern that adheres to this property, the \textit{sandwich transformer}, and show that it improves perplexity on multiple word-level and character-level language modeling benchmarks, at no cost in parameters, memory, or training time.
However, the sandwich reordering pattern does not guarantee performance gains across every task, as we demonstrate on machine translation models.
Instead, we suggest that further exploration of task-specific sublayer reorderings is needed in order to unlock additional gains.\footnote{Our code is available at \url{https://github.com/ofirpress/sandwich_transformer}} 
\end{abstract}

%% file: 01-intro.tex
\section{Introduction}

The transformer layer~\citep{AIAYN} is currently the primary modeling component in natural language processing, playing a lead role in recent innovations such as BERT~\citep{BERT} and GPT-2~\citep{gpt-2}.
Each transformer layer consists of a \textit{self-attention} sublayer ({\Large \texttt{\fs s}}) followed by a \textit{feedforward} sublayer  ({\Large \texttt{\fs f}}), creating an interleaving pattern of self-attention and feedforward sublayers ({\Large \texttt{\fs{sfsfsf}}}\thinspace$\cdots$) throughout a multilayer transformer model.  
To the best of our knowledge, there is no   
reason to expect this particular pattern to be optimal.
We conduct a series of explorations to obtain insights about the nature of transformer orderings that work well, and based on this, we design a new transformer ordering pattern that improves upon the baseline. 

First, we generate random transformer models, varying the number of each type of sublayer, and their ordering, while keeping the number of parameters constant. We train these models on the standard WikiText-103 word-level language modeling benchmark \cite{merity2016pointer},
and observe that some of these random models outperform the original interleaved transformer model, even when the number of self-attention and feedforward layers is not equal.
Our analysis shows that models with more self-attention toward the bottom and more feedforward sublayers toward the top tend to perform better in general.

\begin{figure}[t!] 
\centering
\begin{tabular}{c}
\texttt{\Large\fs{sfsfsfsfsfsfsfsfsfsfsfsfsfsf}} \\
\multirow{2}{*}{\small{(a) Interleaved Transformer}} \\
\\
\\
\texttt{\Large \fs{sssssssfsfsfsfsfsfsfsfffffff}} \\
\multirow{2}{*}{\small{(b) Sandwich Transformer}} \\
\\
\end{tabular}

\caption{A transformer model (a) is composed of interleaved self-attention (green) and feedforward (purple) sublayers. Our sandwich transformer (b), a reordering of the transformer sublayers, performs better on language modeling. Input flows from left to right. }
 \label{fig:sandwichmain}
\end{figure}

Based on this insight, we design a new family of transformer models that follow a distinct sublayer ordering pattern: \emph{sandwich transformers} (Figure~\ref{fig:sandwichmain}).
Our experiments demonstrate that a sandwich transformer outperforms the baseline of~\citet{baevski2018adaptive}.
This result is made more interesting by the fact that our sandwich transformer is simply a reordering of the sublayers in the baseline model, and does not require more parameters, memory, or training time.

Finally, we demonstrate that even though the sandwich transformer is motivated by random search experiments on WikiText-103, it can improve performance on additional domains and tasks. Sandwich transformers achieve state-of-the-art results on the enwik8 character-level language modeling dataset and on an additional word-level corpus, but have no significant effect on machine translation.
We conjecture that tuning transformer reorderings to specific tasks could yield even larger gains, and that further exploration of the ordering space may provide universally beneficial patterns.

%% file: 02-background.tex
\section{Notation}
\label{sec.notation}

\begin{table}[t]
\centering
\input{99tab.1.tex}
\caption{Randomly generated models with 16 self-attention ({\Large \texttt{\fs{s}}}) sublayers and 16 feedforward ({\Large \texttt{\fs{f}}}) sublayers, and their perplexity on the WikiText-103 development set. The baselines (the standard transformer trained with different random seeds) are in bold.}
\label{tab.rs1}
\end{table}

Each transformer layer consists of a self-attention sublayer followed by a feedforward sublayer,
modifying a sequence of vectors $\mathbf{X}_0$ as follows:\footnote{We omit dropout~\cite{dropout} and layer normalization~\cite{layernorm} to simplify the notation.}
\begin{align*}
\mathbf{X}_1 &= \textrm{self-attention}(\mathbf{X}_0) + \mathbf{X}_0\\
\mathbf{X}_2 &= \textrm{feedforward}(\mathbf{X}_1) + \mathbf{X}_1
\end{align*}
Stacking multiple transformer layers creates an interleaved network of sublayers. We denote these models as strings, with {\Large \texttt{\fs{s}}} and {\Large \texttt{\fs{f}}} representing self-attention and feedforward sublayers, respectively. A three-layer transformer network, for example, would be denoted {\Large \texttt{\fs{sfsfsf}}}, with the flow of computation moving from input on the left to output on the right.
Thus, any string in the regular language $(${\Large \texttt{\fs{s}}}$\mid${\Large \texttt{\fs{f}}}$)^\ast$  defines a valid network that uses the same building blocks as the original transformer.  For simplicity, we refer to these alternatives as transformers as well.

%% file: 99tab.1.tex
\small
\begin{tabular}{@{}ll@{}}
\toprule
\textbf{Model}                              & \textbf{PPL}   \\ \midrule
\texttt{ \fs{fsfsfffsffsfsssffsfssfssssffsffs} }&{ 20.74 }\\
\texttt{ \fs{sfssffsffffssssfsfffsfsffsfssssf }}&{ 20.64 }\\
\texttt{ \fs{fsffssffssssffsssssffsfssfsfffff }}&{ 20.33 }\\
\texttt{ \fs{fsffffffsssfssffsfssffsfsssffsss }}&{ 20.27 }\\
\texttt{ \fs{fssffffffsfsssfffssssfffssssffss }}&{ 19.98 }\\
\texttt{ \fs{sssfssfsffffssfsfsfsssffsfsfffsf }}&{ 19.92 }\\
\texttt{ \fs{fffsfsssfsffsfsffsffsssssffssffs }}&{ 19.69 }\\
\texttt{ \fs{fffsffssffsssfssfsssfffffsfsssfs }}&{ 19.54          } \\
\texttt{ \fs{sfsfsfsfsfsfsfsfsfsfsfsfsfsfsfsf} }& { \textbf{19.13}} \\
\texttt{ \fs{fsffssfssfffssssfffsssffffsfssfs }}& { 19.08         } \\
\texttt{ \fs{sfsffssssffssffffsssffsssfsffsff }}& { 18.90         } \\
\texttt{ \fs{sfsfsfsfsfsfsfsfsfsfsfsfsfsfsfsf }}& { \textbf{18.83}} \\
\texttt{ \fs{sssssssffsffsfsfsffffsfffsfssffs }}& { 18.83         } \\
\texttt{ \fs{sffsfsffsfsssffssfssssssfffffffs }}& { 18.77         } \\
\texttt{ \fs{sssfssffsfssfsffsfffssffsfsffssf }}& { 18.68         }   \\
\texttt{ \fs{fffsssssfffsfssssffsfsfsfssffsff }}& { 18.64         }  \\
\texttt{ \fs{sfffsssfsfssfsssssfssfffffsfffsf }}& { 18.61         }  \\
\texttt{ \fs{ssffssfssssffffffssffsssfsffssff }}& { 18.60         }   \\
\texttt{ \fs{fsfsssssfsfsfffffsfffsffssffssss }}& { 18.55         }  \\
\texttt{ \fs{sfsfsfsfsfsfsfsfsfsfsfsfsfsfsfsf} }& { \textbf{18.54}} \\
\texttt{ \fs{sfsfsfsfsfsfsfsfsfsfsfsfsfsfsfsf}}&  { \textbf{18.49}} \\
\texttt{ \fs{fsfsssssfsfffssfsffsfsfsfsffffss }}& { 18.38         }  \\
\texttt{ \fs{sfssffsfsfsffsssssfffsssfffsffsf }}& { 18.28         }  \\
\texttt{ \fs{sfsfsfsfsfsfsfsfsfsfsfsfsfsfsfsf} }& { \textbf{18.25}} \\
\texttt{ \fs{sfsfssfsssffsfsfsfsffffssffsfssf }}& { 18.19         }  \\
\bottomrule
\end{tabular}

%% file: 03-hypothesis.tex
\begin{figure}[t]
\centering
\includegraphics[width=0.48\textwidth]{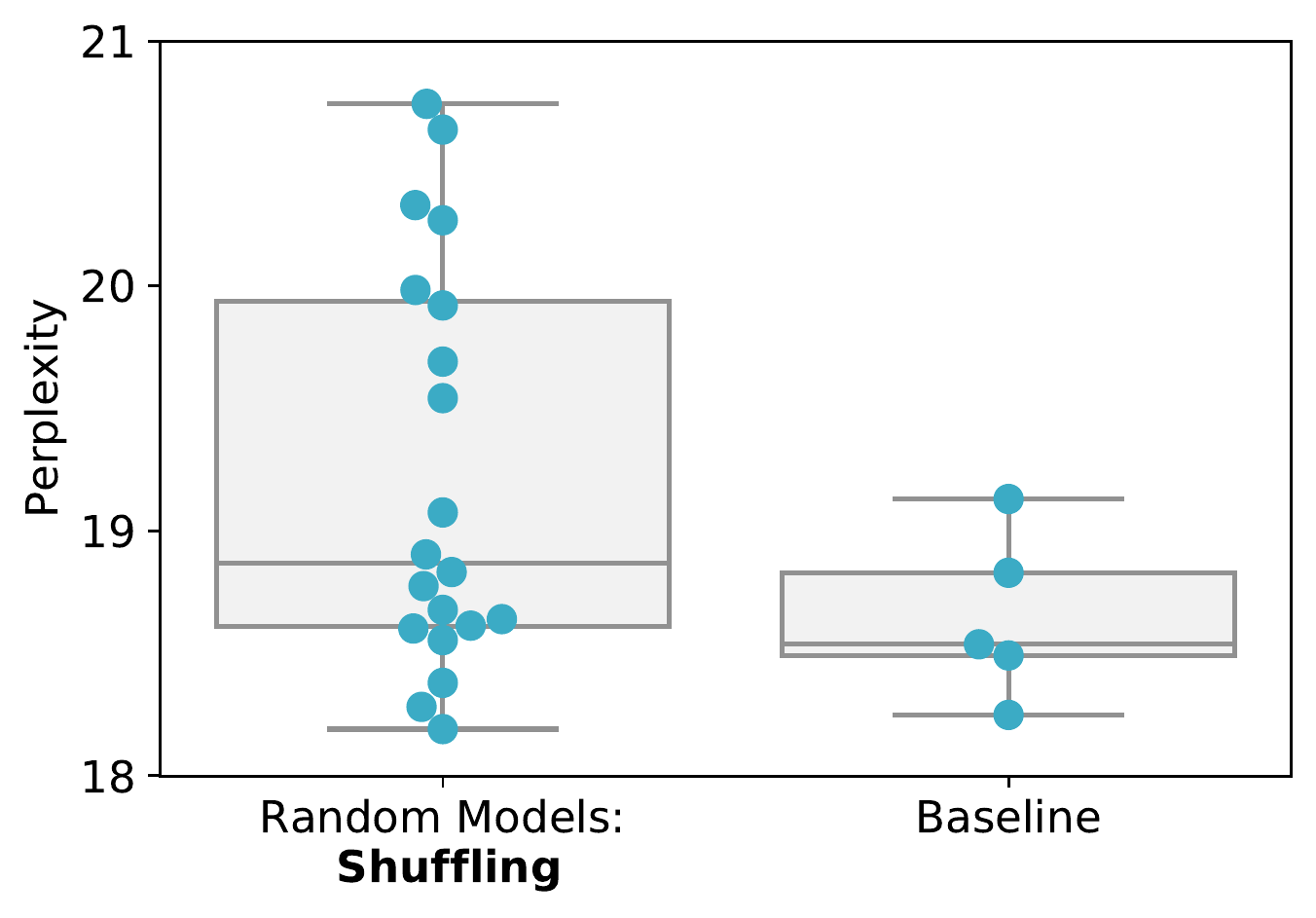}
\caption{The perplexities on the WikiText-103 development set of 20 randomly generated models with 16 self-attention and 16 feedforward sublayers and of the 5 baselines (the standard transformer trained with different random seeds). }
\label{fig.shuffle}
\end{figure}

\section{Random Search} \label{sec.exploration}

We conduct a series of experiments to understand which transformer networks work well and whether particular architectural patterns can  improve performance.
First, we generate random transformer models while keeping the number of parameters constant.
We then train these random models to determine whether the interleaving pattern ({\Large \texttt{\fs{sfsfsf}}}\thinspace$\cdots$) is optimal (Section~\ref{sec.interleaving}), and whether balancing the number of self-attention and feedforward sublayers is desirable (Section~\ref{sec.f_eq_s}).
Finally, we analyze additional properties of these random models, and find that those with more self-attention at the beginning and more feedforward sublayers near the end tend to outperform the standard interleaved model (Section~\ref{sec.analysis}).

\paragraph{Experimental Setup}
Our baseline is the strong transformer language model of~\citet{baevski2018adaptive}, trained on WikiText-103 \citep{merity2016pointer}. WikiText-103 contains roughly 103 million tokens from English Wikipedia, split into train, development, and test sets by article.  The Baevski and Auli model contains $16$ transformer layers of $d=1024$ dimensions, with $16$ heads in each self-attention sublayer, and feedforward sublayers with an inner dimension of $4096$.
In this setting, each self-attention sublayer contains $4d^2$ parameters, while each feedforward sublayer contains $8d^2$ parameters (excluding bias terms, which have a marginal contribution).
Thus, each {\Large \texttt{\fs{f}}} sublayer contains twice the parameters of a {\Large \texttt{\fs{s}}} sublayer, following the parameter ratio between self-attention and feedforward sublayers described in~\citet{AIAYN}.

All of our experiments use the same hyperparameters as Baevski and Auli's original model.
To set an accurate baseline, we train the baseline model (the standard interleaved transformer) with five different random seeds, achieving 18.65 $\pm$ 0.24 perplexity on the development set.

\begin{table}[t]
\centering
\input{99tab.2.tex}

\caption{Randomly generated models with the same number of parameters as the baseline, and their perplexity on the WikiText-103 development set. The baselines (the standard transformer trained with different random seeds) are in bold.}
\label{tab.budget}
\end{table}

\begin{figure}[t]
\centering
\includegraphics[width=0.48\textwidth]{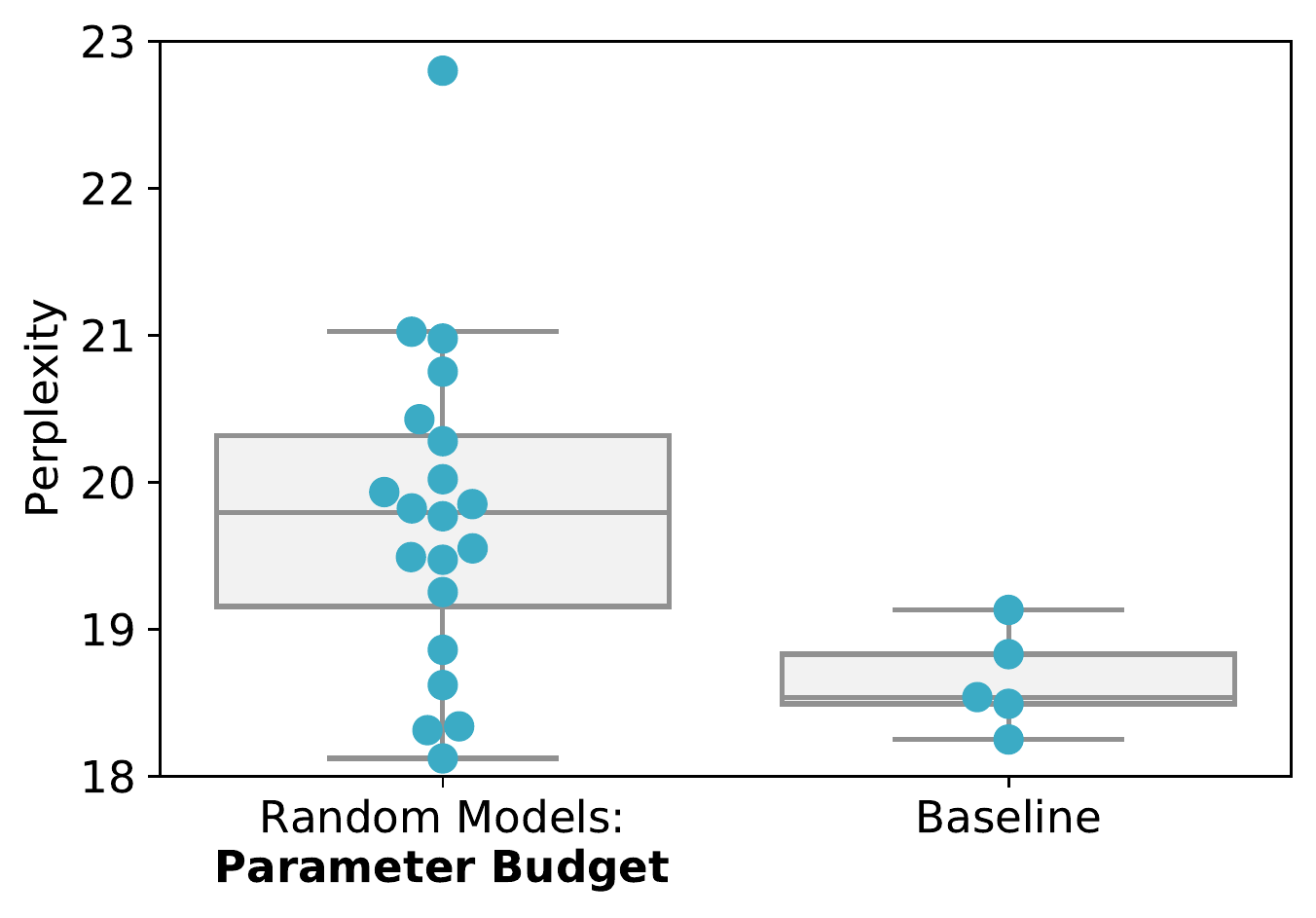}
\caption{The perplexities on the WikiText-103 development set of 20 randomly generated models with the same number of parameters as the baseline, and of the 5 baselines (the standard transformer trained with different random seeds).} 
\label{fig.budget}
\end{figure}

\subsection{Is Interleaving Optimal?}
\label{sec.interleaving}

In the baseline 16-layer transformer model, 16 sublayers of each type are interleaved. Can we improve model performance by simply rearranging them?
We thus generate 20 random transformer models with 16 self-attention sublayers and 16 feedforward sublayers, randomly permuted, and train these models from scratch, without modifying any of the hyperparameters. 
Table~\ref{tab.rs1} shows the entire sample, while Figure~\ref{fig.shuffle} plots the perplexity distributions of the shuffled transformers and the baseline side by side.

We observe that 7 of the 20 randomly-permuted models perform at least as well as the interleaved baseline's average performance, with the best model achieving $18.19$ perplexity.
While the average performance of the baseline model beats the average performance of these random models, the fact that a third of our random models outperformed the average baseline suggests that a better ordering than interleaving probably exists.

\begin{figure*}[t]
\begin{subfigure}[t]{0.48\linewidth}
\centering
\includegraphics[width=\textwidth]{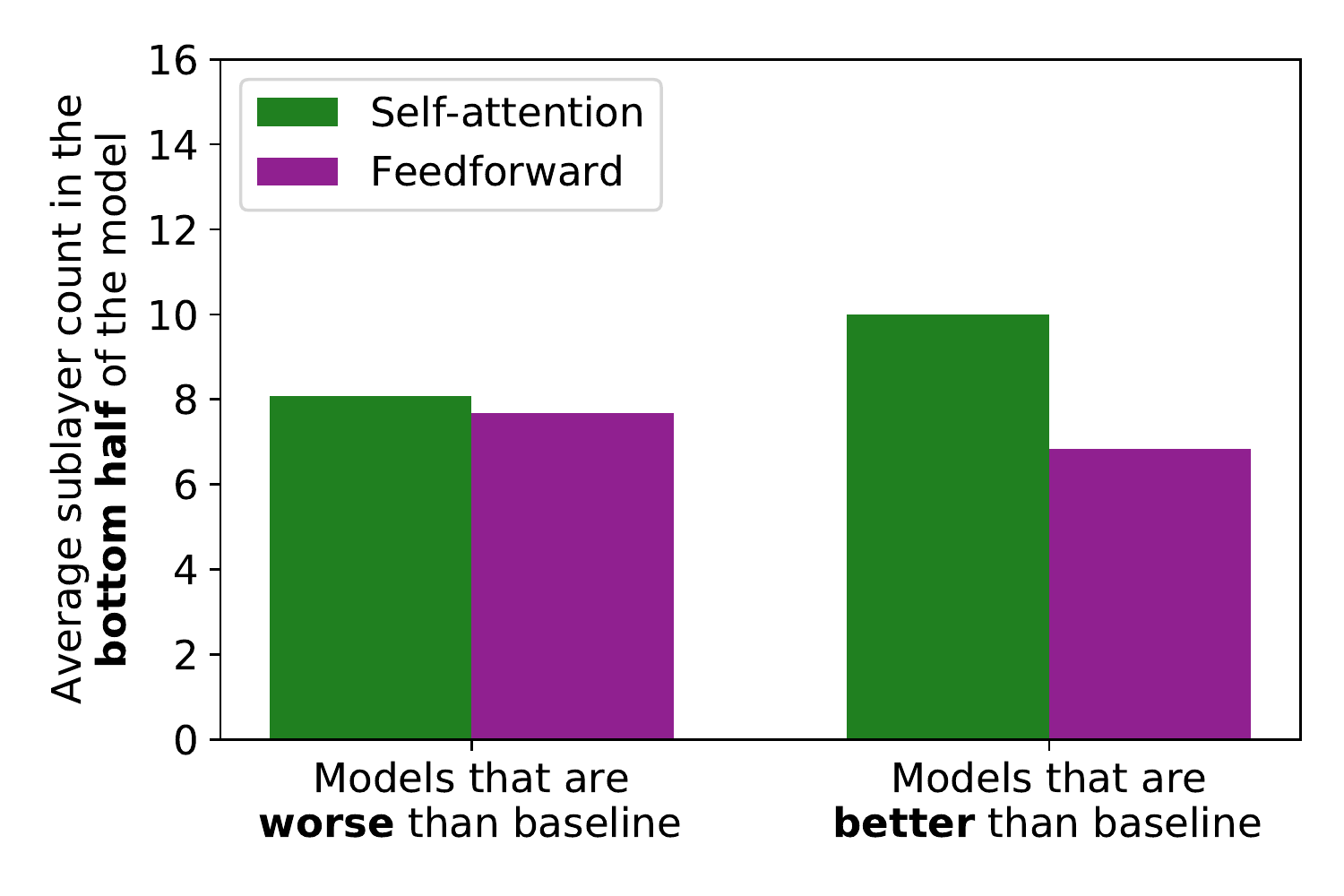}
\caption{}
\end{subfigure}
~
\begin{subfigure}[t]{0.48\linewidth}
\centering
\includegraphics[width=\textwidth]{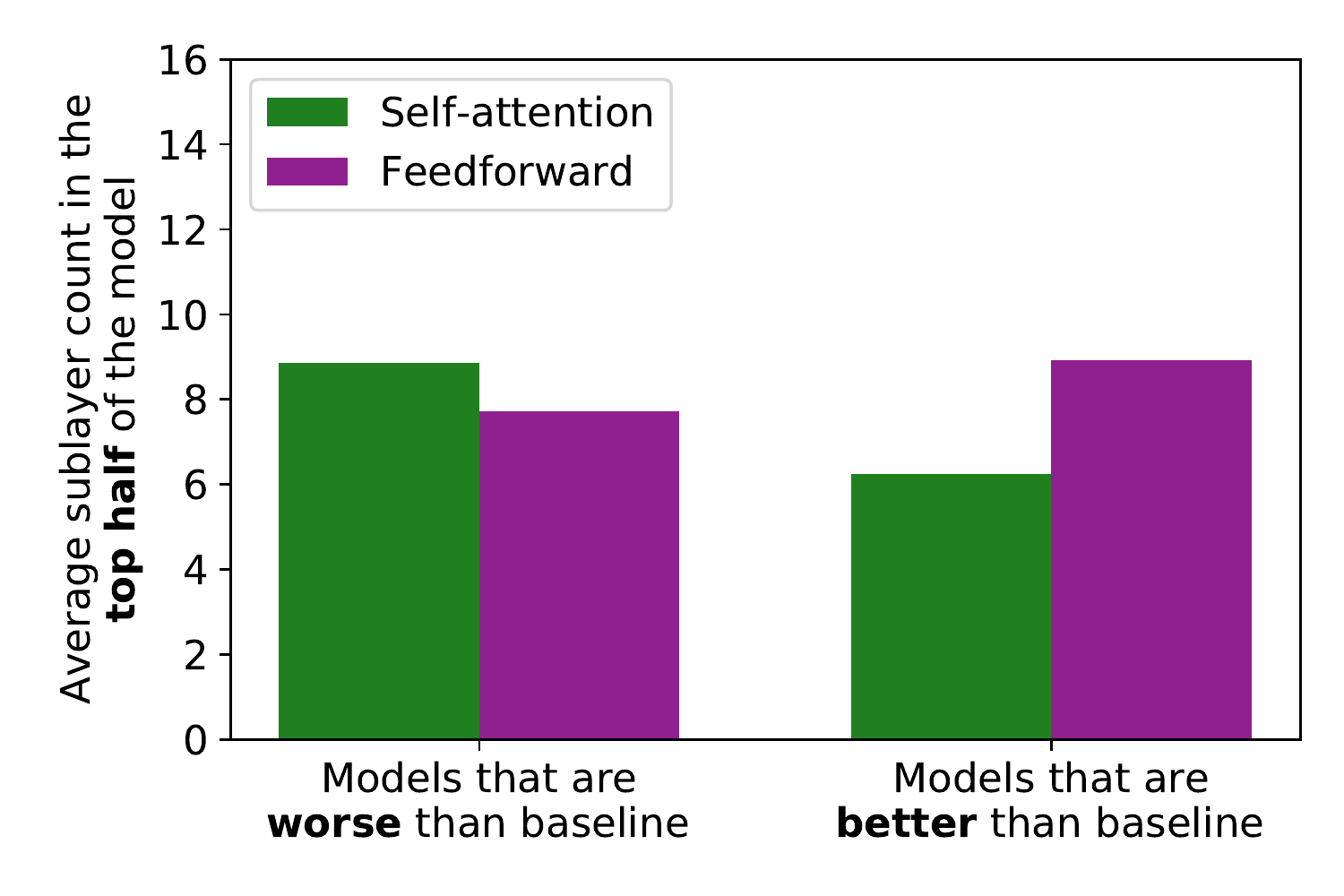}
\caption{}
\end{subfigure}
\caption{Analysis of sublayer distribution in models that do better or worse than the average  baseline, split across bottom (a) and top (b) halves of the model.}
\label{fig.split}
\end{figure*}

\subsection{Are Balanced Architectures Better?}
\label{sec.f_eq_s}

Is it necessary to have an identical number of sublayers of each type, or could models with more self-attention (or more feedforward) sublayers yield better results?
To find out, we generate 20 unbalanced transformer models by randomly selecting one sublayer at a time (either {\Large \texttt{\fs{s}}} or {\Large \texttt{\fs{f}}} with equal probability) until the parameter budget is exhausted. Since a feedforward sublayer contains double the parameters of a self-attention sublayer, the networks' depth is not necessarily 32 sublayers as before and can range from 24 (all {\Large \texttt{\fs{f}}}) to 48 (all {\Large \texttt{\fs{s}}}).
Table~\ref{tab.budget} shows the entire sample, while Figure~\ref{fig.budget} plots the perplexity distributions of the randomly-generated transformers and the baseline side by side.

We see that four of the generated unbalanced models outperform the average baseline transformer.
The best performing random model reaches a perplexity of 18.12 and has 12 self-attention and 18 feedforward sublayers.
Both the average and the median perplexities of this sample of unbalanced models are worse than those of the balanced permuted models (Section~\ref{sec.interleaving}).
We do not observe any preference for more sublayers of one type over the other; there are self-attention-heavy and feedforward-heavy models in both the top five and the bottom five of the results table.
While offering no guarantees -- given the small sample sizes and fixed hyperparameters -- we conclude that a balanced number of self-attention and feedforward sublayers seems to be a desirable property, though not a necessary one.

\subsection{Attention First, Feedforward Later}
\label{sec.analysis}

So far, it is not clear which characteristics make one transformer model more successful than another; for example, measuring the number of times each sublayer type appears in the network does not reveal any strong correlation with performance.
However, analyzing the bottom (or top) half of the network in isolation reveals an interesting property.

We first split the models to those that perform better than the average baseline and those that do not.
We then slice each one of the previously-generated random models in half by parameter count (e.g., {\Large \texttt{\fs{ssssff}}} would be split to {\Large \texttt{\fs{ssss}}} and {\Large \texttt{\fs{ff}}}, since every {\Large \texttt{\fs{f}}} contains twice as many parameters as an {\Large \texttt{\fs{s}}}), and count how many sublayers of each type appear in each slice.

Figure~\ref{fig.split} shows that models that outperform the average baseline tend to have more self-attention {\Large \texttt{\fs{s}}} in the first (bottom) half of the network and more {\Large \texttt{\fs{f}}} in the second (top) half.
While we do not have a good hypothesis to \emph{explain} this phenomenon, we can \emph{exploit} it to improve transformers (Section~\ref{sec.designing}).

%% file: 99tab.2.tex
\small
\begin{tabular}{@{}ll@{}}
\toprule
\textbf{Model}                              & \textbf{PPL}   \\ \midrule
\texttt{  \fs{sfffssfsfsfssffffsfsffsffffff          } }&{ 22.80          } \\
\texttt{  \fs{sffssfsssssssssssssfsfsssfsffsssfsssfs } }&{ 21.02          } \\
\texttt{  \fs{ssssssffsffffssfffffsssfsfsssssssss    } }&{ 20.98          } \\
\texttt{  \fs{fffffffffsffssffsffssssfsfsssf         } }&{ 20.75          } \\
\texttt{  \fs{fssfsssffffffssfsssfsfffssssfsfss      } }&{ 20.43          } \\
\texttt{  \fs{sffsffffffsfsfssfsssfsfsfssfssfs       } }&{ 20.28          } \\
\texttt{  \fs{sffssffsfffsfsfssssffffffssssff        } }&{ 20.02          } \\
\texttt{  \fs{fsffsfssffffsfsfffsfffssfffsss         } }&{ 19.93          } \\
\texttt{  \fs{sffsffssffsfsffsssfsssssfsssfffsss     } }&{ 19.85          } \\
\texttt{  \fs{ssfffffffssfffssfssffsfsfsffsf         } }&{ 19.82          } \\
\texttt{  \fs{sfsfsfffsfffssfsfffsffssfsfsfss        } }&{ 19.77          } \\
\texttt{  \fs{sfsffsssffsffsssfssfffffssssfsssf      } }&{ 19.55          } \\
\texttt{  \fs{sffsfssfffsffsfssssfsfsffffsfsss       } }&{ 19.49          } \\
\texttt{  \fs{sffffsffssssfsssfssfffsssfssssfsfs     } }&{ 19.47          } \\
\texttt{  \fs{fsssffssssssfsfsfsffsffffssfsfssss     } }&{ 19.25          } \\
\texttt{  \fs{sfsfsfsfsfsfsfsfsfsfsfsfsfsfsfsf       } }&{ \textbf{19.13} } \\
\texttt{  \fs{fssssssfsfsfsfffsfsssfssffssssfsff     } }&{ 18.86          } \\
\texttt{  \fs{sfsfsfsfsfsfsfsfsfsfsfsfsfsfsfsf       } }&{ \textbf{18.83} } \\
\texttt{  \fs{ssfsfsssfsssssffsfsfsssfssfsfsssssssf  } }&{ 18.62          } \\
\texttt{  \fs{sfsfsfsfsfsfsfsfsfsfsfsfsfsfsfsf       } }&{ \textbf{18.54} } \\
\texttt{  \fs{sfsfsfsfsfsfsfsfsfsfsfsfsfsfsfsf       } }&{ \textbf{18.49} } \\
\texttt{  \fs{sssfsffsfssfsssffsffffffssfsfff        } }&{ 18.34          } \\
\texttt{  \fs{sssfsfsffsssfsfffffsfsffffsssff        } }&{ 18.31          } \\
\texttt{  \fs{sfsfsfsfsfsfsfsfsfsfsfsfsfsfsfsf       }} &{ \textbf{18.25} } \\
\texttt{  \fs{ssssssfsssffffsfsfffffffffffsf         }} &{ 18.12          } \\
\bottomrule
\end{tabular}

%% file: 04-design.tex
\section{Designing a Better Transformer} \label{sec.designing}

Our analysis in the previous section motivates designing a transformer model that is heavy on self-attention at the bottom and feedforward sublayers at the top, while at the same time containing a more-or-less balanced amount of both sublayer types.
As a first attempt to manually design a better transformer, we take this hypothesis to the extreme, and train a transformer model of 16 self-attention sublayers followed by 16 feedforward sublayers ({\Large \texttt{\fs{s}}}$^{16}${\Large \texttt{\fs{f}}}$^{16}$). This model achieves 18.82 perplexity, which is comparable to the performance of the baseline with the same number of parameters.

We next generalize this model and the original interleaved transformer, creating the family of \emph{sandwich transformers}.
A sandwich$^n_k$ transformer consists of $2n$ sublayers in total ($n$ of each type), conforming to the regular expression {\Large \texttt{\fs{s}}}$^k(${\Large \texttt{\fs{sf}}}$)^{n-k}$\thinspace{\Large \texttt{\fs{f}}}$^k$. The first $k$ sublayers are purely self-attention ({\Large \texttt{\fs{s}}}), while the last $k$ are feedforward sublayers ({\Large \texttt{\fs{f}}}). In between, we use the original interleaving pattern ({\Large \texttt{\fs{sf}}}) to fill the remaining $2(n-k)$ sublayers. When $k=0$, we get the original transformer model, and when $k=n-1$ (its maximal value) we get the previously mentioned {\Large \texttt{\fs{s}}}$^{n}${\Large \texttt{\fs{f}}}$^{n}$ model. We refer to $k$ as the transformer's \emph{sandwich coefficient}. 

\begin{figure}[t]
\centering
\includegraphics[width=0.48\textwidth]{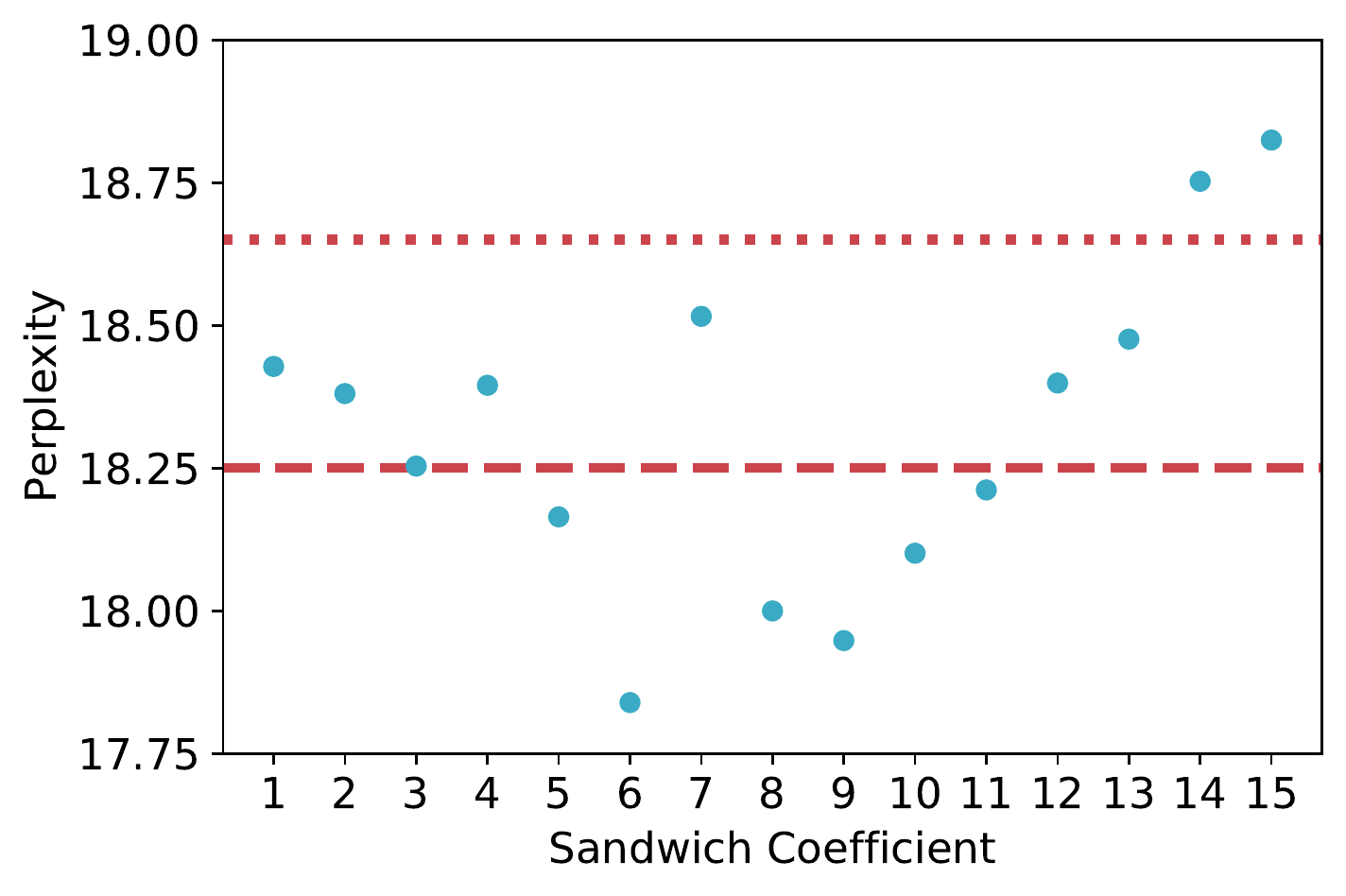}
\caption{The transformer's sandwich coefficient ($k$) and validation perplexity, for  $k \in \{1,\dots,15\}$. The dotted line is the average baseline model's perplexity (trained with different random seeds), whereas the dashed line represents the best baseline model.} 
\label{fig.sandwichperplexity}
\end{figure}

We train sandwich transformers for $n=16$ (to remain within the same parameter budget as our baseline language model) and all values of $k \in \{0, \ldots ,15\}$. Figure~\ref{fig.sandwichperplexity} shows the transformer's performance as a function of the sandwich coefficient $k$. 
With the exception of $k=14,15$, all sandwich transformers achieve lower perplexities than the average baseline transformer.
Of those, 6 models outperform the best baseline transformer ($k=5,6,8,9,10,11$).
The best performance of 17.84 perplexity is obtained when $k = 6$.
We compare this model to the baseline on WikiText-103's test set.

Table~\ref{tab.sandwichtest} shows that, despite its simple design, the sandwich transformer outperforms the original transformer baseline by roughly double the gap between the baseline \cite{baevski2018adaptive} and Transformer XL \cite{transformerXL}.
This improvement comes at no extra cost in parameters, data, memory, or computation; we did not even change any of the original hyperparameters, including the number of training epochs. 

\begin{figure}[h]
\centering
\includegraphics[width=0.45\textwidth]{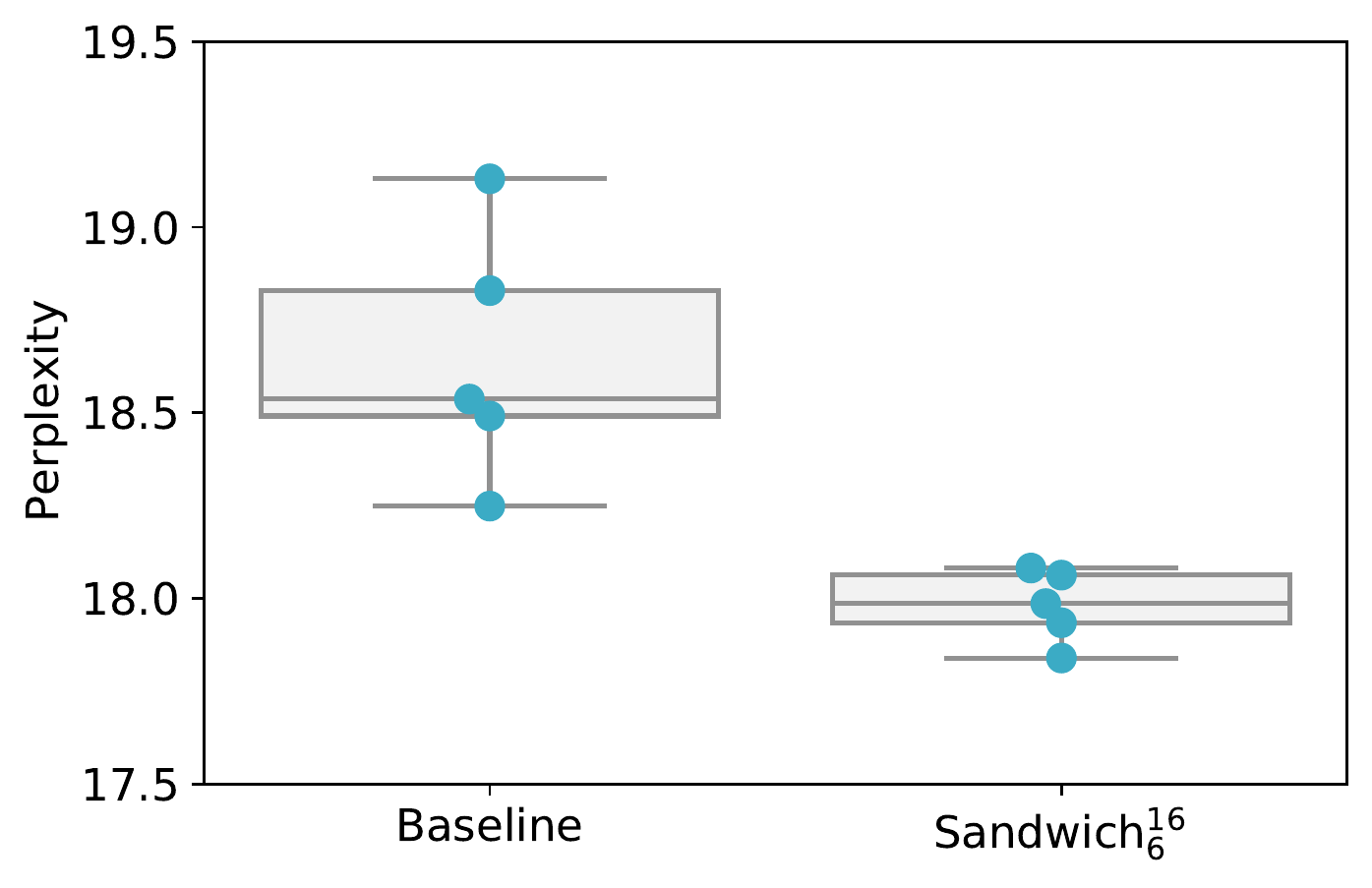}
\caption{Performance on the WikiText-103 development set of the Sandwich$^{16}_6$ transformer and the baseline. Each model is trained with 5 different random seeds to assess the perplexity distribution.}
\label{fig:sandwichdist}
\end{figure}

To check whether this advantage is consistent, we train 4 more sandwich$_6^{16}$ models with different random seeds (5 in total) and evaluate them on the development set, to avoid evaluating our model more than once on the test set. This is the only experiment in which we modify our model's random seed.
Figure~\ref{fig:sandwichdist} shows that we obtain a mean perplexity value of 17.98 with a standard deviation of 0.10, while the baseline achieves 18.65 mean perplexity, with a larger standard deviation of 0.34 (these values reflect \emph{development} set performance, not test set performance as in Table~\ref{tab.sandwichtest}).

In very recent work, kNN-LM \cite{urvashi} set a new state of the art on WikiText-103, surpassing other recent models by a wide margin. The model achieves this result by storing the entire training set in an auxiliary memory component. Since this approach appears orthogonal to ours, it is quite possible that kNN-LM could benefit from sublayer reordering as well.

\begin{table}[t]
\centering
\small
\begin{tabular}{@{}ll@{}}
\toprule
\textbf{Model}   
& \textbf{Test}   \\
\midrule
Baseline \cite{baevski2018adaptive} & 18.70 \\
Transformer XL \cite{transformerXL} & 18.30 \\
kNN-LM \cite{urvashi} & 15.79 \\
\midrule
Baseline (5 Runs)  
& 18.63 $\pm$ 0.26 \\
Sandwich$^{16}_6$ 
& 17.96 \\
\bottomrule
\end{tabular}
\caption{Performance on the WikiText-103 test set. We compare the best sandwich transformer to the unmodified, interleaved transformer baseline \cite{baevski2018adaptive} trained over 5 random seeds and to other previously reported results.}
\label{tab.sandwichtest}
\end{table}

%% file: 05-beyond.tex
\section{One Reordering to Rule Them All?}

The sandwich transformer is a manually-crafted pattern motivated by the performance of random sublayer reorderings of the \citet{baevski2018adaptive} model, trained on the WikiText-103 word-level language modeling benchmark \cite{merity2016pointer}.

Does this particular pattern improve performance in other settings as well?
To find out, we apply sandwich transformers to three other tasks:  word-level language modeling on a different domain (Section~\ref{sec.books}), character-level language modeling (Section~\ref{sec.charlm}), and machine translation (Section~\ref{sec.nmt}). 

Results show that as we drift away from our original setting, sandwich transformers provide diminishing gains, but always perform at least as well as the baseline transformers (provided that the sandwich coefficient is properly tuned).
This finding suggests that different settings may benefit from different sublayer reordering patterns.

\subsection{Books-Domain Language Modeling}
\label{sec.books}

\begin{table}[t]
\centering
\small
\begin{tabular}{@{}ll@{}}
\toprule
\textbf{Model} & \textbf{PPL}  \\
\midrule
Baseline (5 runs) & 11.89 $\pm$ 0.35 \\
kNN-LM \cite{urvashi} & 10.89 \\ 
Sandwich$^{16}_7$ & 10.83 \\
\bottomrule
\end{tabular}
\caption{Performance on the Toronto Books Corpus language modeling test set. The baseline model \cite{baevski2018adaptive} is trained over 5 random seeds. The sandwich coefficient is tuned on the validation set and we run our model on the test set only once.}
\label{tab.bookstest}
\end{table}

\begin{table*}[t]
\centering
\small
\begin{tabular}{@{}lll@{}}
\toprule
\textbf{Model} & \textbf{text8 (BPC)} & \textbf{enwik8 (BPC)}  \\
\midrule
Transformer-XL \cite{transformerXL} & 1.08 & 0.99 \\ 
Adaptive Span \cite{Sukhbaatar2019} & 1.07 & 0.98 \\
Compressive \cite{compressive} & --- & 0.97 \\
\midrule
Baseline (Adaptive Span; 5 Runs) & 1.0802 $\pm$ 0.0103 & 0.9752 $\pm$ 0.0008 \\
Sandwich$^{24}_3$ & 1.076 & --- \\
Sandwich$^{24}_5$ & --- & 0.968 \\
\bottomrule
\end{tabular}
\caption{Performance on character-level language modeling, evaluated on the enwik8 and text8 test sets. The baseline model \cite{Sukhbaatar2019} is trained over 5 random seeds. The sandwich coefficient is tuned on each benchmark's validation set, and we run our model on the test only once.}
\label{tab.charlm}
\end{table*}

We first apply sandwich transformers to a different domain, while retaining the other architectural aspects and hyperparameter settings from \citet{baevski2018adaptive}.
Specifically, we use the Toronto Books Corpus  \cite{moviebook}, which has previously been used to train GPT \cite{gpt-1} and  also  BERT \cite{BERT} (combined with Wikipedia). The corpus contains roughly 700M tokens.

We use the same train/validation/test split as~\citet{urvashi}, as well as their tokenization, which uses BERT's vocabulary of 29K byte-pair encodings.
Since the vocabulary is much smaller than WikiText-103's, we replace the adaptive word embedding and softmax of \citet{baevski2018adaptive} with a tied word embedding and softmax matrix \cite{press2017,inan2017}.
Finally, we tune the sandwich coefficient on the development set for $k \in \{ 4, \ldots, 8\}$, i.e., a neighborhood of 2 around the best value we found for WikiText-103 ($k=6$).

Table~\ref{tab.bookstest} shows that the sandwich transformer transfers well to the books domain, improving performance by 1.06 perplexity, achieving similar performance to the datastore-augmented kNN-LM \cite{urvashi}, which is the state of the art on WikiText-103 (see Section~\ref{sec.designing}).

\subsection{Character-level Language Modeling}
\label{sec.charlm}

Modeling text as a stream of characters, rather than word or subword tokens, presents a different modeling challenge: long-range dependencies become critical, and the vocabulary takes on a more uniform distribution.
We apply our sandwich reordering to the adaptive span model of \citet{Sukhbaatar2019}, which is state of the art on the popular English-language benchmark text8 and is currently a close second on enwik8.\footnote{{Both datasets are taken from \url{http://mattmahoney.net/dc/textdata.html}}}
The adaptive span model learns to control each attention head's maximal attention span, freeing up memory in the bottom layers (which typically need very short attention spans) and applying it to the top layers, allowing the top-level attention heads to reach significantly longer distances.
The adaptive span model's efficient use of attention also results in a significant speed boost.

We tune the sandwich coefficient on the development set for $k \in \{ 1, \ldots, 8\}$ (the baseline model has 24 transformer layers). We do not modify any hyperparameters, including the number of training epochs.
Table~\ref{tab.charlm} compares the baseline model's performance with the sandwich transformer's. 
On text8, the sandwich transformer performs within the baseline's random seed variance. On enwik8, the sandwich transformer gains an improvement of about 0.007 bits-per-character, matching the state of the art results obtained by the Transformer-XL-based Compressive Transformer of~\citet{compressive}.

However, our approach is able to achieve this result without applying the Transformer-XL's recurrent attention, which is much slower~\citep{Sukhbaatar2019}, and without adding additional parameters (the compressive transformer uses 277M parameters, while our baseline and sandwich models use only 209M).

\subsection{Machine Translation}
\label{sec.nmt}

\paragraph{Sandwich Decoders}
Tranformer-based translation models~\cite{AIAYN} consist of an encoder and decoder, where the encoder has interleaved self-attention and feedforward sublayers (just as in language models), while the decoder includes an additional sublayer, cross-attention ({\Large \texttt{\fs{c}}}), between every pair of self-attention and feedforward sublayers. Cross-attention sublayers attend to the encoder's representations of the input sentence's tokens.

Following our notation from Section~\ref{sec.notation}, a transformer decoder layer modifies the sequence of tokens in the target language $\mathbf{Y_0}$, using the encoded source tokens $\mathbf{X}$, as follows:
\begin{align*}
\mathbf{Y}_1 &= \textrm{self-attention}(\mathbf{Y}_0) + \mathbf{Y}_0\\
\mathbf{Y}_2 &= \textrm{cross-attention}(\mathbf{Y}_1, \mathbf{X}) + \mathbf{Y}_1\\
\mathbf{Y}_3 &= \textrm{feedforward}(\mathbf{Y}_2) + \mathbf{Y}_2
\end{align*}

Applying the sandwich pattern to the encoder follows the same methodology as our previous experiments. However, for the decoder, we group the self-attention ({\Large \texttt{\fs{s}}}) and cross-attention ({\Large \texttt{\fs{c}}}) sublayers, and treat them as a single unit for reordering purposes ({\Large \texttt{\fs{sc}}}). For example, a three layer decoder ({\Large \texttt{\fs{scfscfscf}}}) with a sandwiching coefficient of $k=1$ would be: {\Large \texttt{\fs{scscfscff}}}.
We apply the sandwich pattern to either the encoder or decoder separately, while keeping the other stack in its original interleaved pattern.

\paragraph{Experiment Setting}
As a baseline, we use the large transformer model (6 encoder/decoder layers, embedding size of 1024, feedforward inner dimension of 4096, and 16 attention heads) with the hyperparameters of \citet{scalingnmt}. We also follow their setup for training and evaluation: we train on the WMT 2014 En-De dataset which contains 4.5M sentence pairs; we validate on newstest13 and test on newstest14. We use a vocabulary of 32K symbols based on a joint source and target byte pair encoding~\cite{bpe}. For inference we use beam search with a beam width of 4 and length penalty of 0.6, following~\citet{AIAYN} and \citet{scalingnmt}. As before, we do not modify our model's hyperparameters or training procedure.

\paragraph{Results}
Table~\ref{tab.allnmt} shows that reordering of either the encoder or decoder does not have a significant impact on performance, across the board.
We also find that using the most extreme sandwich decoder ({\Large \texttt{\fs{s}}}{\Large \texttt{\fs{c}}})$^{6}${\Large \texttt{\fs{f}}}$^{6}$ performs almost exactly the same as the average baseline; this result is consistent with our observation from Section~\ref{sec.designing}, where we show that the extreme sandwich language model ({\Large \texttt{\fs{s}}}$^{16}${\Large \texttt{\fs{f}}}$^{16}$) performs as well as the baseline.

\begin{table}[!t]
\centering
\small

\begin{tabular}{lcc}
\toprule
\textbf{Sandwich} & \textbf{Encoder} &  \textbf{Decoder} \\
\textbf{Coefficient} & \textbf{Sandwich} &  \textbf{Sandwich } \\
\midrule
0 (Baseline)                                                                  & \multicolumn{2}{c}{28.74 $\pm$ 0.15}  \\ 
\midrule
1                                                                  &  28.71       & 28.64     \\ 
2                                                                  &  28.71       & 28.56     \\ 
3                                                                  &  28.81       & 28.67     \\ 
4                                                                  &  28.48       & 28.66     \\ 
5                                                                  &  28.45       & 28.76     \\ \bottomrule
\end{tabular}

\caption{BLEU on newstest2014 En-De. Our encoder (decoder) sandwich model keeps the decoder (encoder) unmodified. We train the baseline model (Transformer-large with the hyperparameters of \citealp{scalingnmt}) 5 times with different random seeds.}
\label{tab.allnmt}
\end{table}

\paragraph{Discussion}
This experiment indicates that a reordering pattern that benefits one particular task (language modeling) might not carry the same performance gains to another (machine translation).
However, it also demonstrates the general robustness of transformer architectures to sublayer reordering, as we did not observe any major performance degradation. 
Since the sandwich pattern naively groups self- and cross-attention sublayers together, it is also possible that a reordering pattern that takes all three sublayer types into account could potentially improve performance.

%% file: 06-analysis.tex
\section{Analysis}

At the time of writing, we do not have an explanation for why sublayer reordering improves performance on language modeling.
However, we are able to determine that sandwich transformers spread their attention in a \emph{different} fashion than interleaved models.

We analyze two baseline models and two sandwich$^{16}_6$ models trained with different seeds on the WikiText-103 dataset, by first recording the attention values that each token's heads assign to all other tokens during inference on the validation set. 
Given the attention outputs of two models, we then compute the models' \emph{attention distance} for each token, and for each self-attention sublayer. This metric compares the attention distribution in the $i$th self-attention sublayer of the first model to that of the $i$th self-attention sublayer of the second model, for a specific token. 

Given a token and a self-attention sublayer, we use the Hungarian algorithm~\cite{kuhn1955hungarian} to find a matching of heads in the first model to heads in the second model $[a_1, b_1], \dots, [a_8, b_8]$ such that $\sum_{i=1}^8 \text{EMD}(a_i, b_i)$ is minimized, where $\text{EMD}(a_i, b_i)$ is the earth mover's (Wasserstein) distance between the attention distributions of head $a_i$ in the first model and head $b_i$ in the second model. That minimal value is the attention distance for that token, in that layer.
We then average the attention distances across all tokens and layers.

Table~\ref{tab.analysis} shows the average attention distances between every pair of models. We observe that models of the same architecture have significantly lower attention distances than models with different sublayer orderings.
This indicates that sublayer reordering has a strong effect on the attention function that the model learns in each head.
Future investigations of what this difference is, in a qualitative sense, could potentially provide important insights for designing better reordering patterns.

\begin{table}[t]
\centering
\small
\begin{tabular}{@{}ll@{}}
\toprule
\textbf{Model Pair} & \textbf{Average Attention Distance} \\
\midrule
Baseline -- Baseline & $1.081 \cdot 10^{-3}$ \\
Sandwich  -- Sandwich & $1.067\cdot 10^{-3}$ \\
Baseline -- Sandwich & $1.289 \cdot 10^{-3} \pm 0.049 \cdot  10^{-3}$\\
\bottomrule
\end{tabular}
\caption{The average attention distance, on the WikiText-103 validation dataset, of each model pair. Since there are two baselines and two sandwich transformers (initialized with different random seeds), the distance between the baseline and sandwich models is averaged over all four baseline-sandwich combinations. }
\label{tab.analysis}
\end{table}

%% file: 07-related.tex
\section{Related Work}

\subsection{Neural Architecture Search}
In this paper, we manually search through a constrained transformer architecture space, after analyzing the results of two small-scale random searches. This human-in-the-loop method for architecture search has advantages over previous methods~\cite{jozefowicz2015empirical,zoph2016neural,efficientnet} since it requires that only a few dozen models be trained, unlike typical architecture search methods that require training thousands of instances, consuming massive computational resources.

While we do find a better performing transformer, our goal is not only to do so, but to better understand how sublayer ordering affects transformer models. Future work could apply methods from the architecture space literature to the sublayer ordering problem. Furthermore, a better understanding of the inner workings of transformers could inspire more efficient, constrained architecture search.

\subsection{Transformer Modifications}
Much recent work has been devoted to improving transformers by modifying their sublayers. This includes sparsifying their attention patterns, either in an input-based manner (as in~\citealp{correia2019adaptively}), or in a static manner (as in~\citealp{startransformer}). \citet{evolvedtransformer} proposed modifying the transformer by adding convolutions and changing the activation function, while others have demonstrated that different initialization schemes~\cite{zhang2019improving} and repositioning the layer normalization \cite{nguyen2019transformers} can also have a positive effect on performance. 

In this paper, we do not modify the sublayers at all, but simply rearrange their order. The performance gains from sublayer reordering are orthogonal to improving the sublayers themselves, and could be combined to achieve even better performance.

Recently, \citet{lu2019understanding} introduced a new transformer ordering, where instead of stacking layers of the form {\Large \texttt{\fs{sf}}} (as in the vanilla interleaved transformer), they stack layers of the form {\Large \texttt{\fs{fsf}}}. In order keep the total parameter count unchanged, Lu et al.~cut the hidden dimension of their feedforward sublayers by half. However, the overall depth of the network is increased by 50\%, which causes a similar increase in the model's inference time \cite{sanh_2019}. 

%% file: 08-conclusion.tex
\section{Conclusion}
We train random transformer models with reordered sublayers, and find that some perform better than the baseline interleaved transformer in language modeling. We observe that, on average, better models contain more self-attention sublayers at the bottom and more feedforward sublayer at the top. This leads us to design a new transformer stack, the sandwich transformer, which significantly improves performance over the baseline at no cost in parameters, memory, or runtime. 

We then show that the sandwich ordering also improves language modeling performance on a different word-level language modeling benchmark, and that the sandwich pattern can be used to achieve state of the art results on character-level language modeling. 
Although sandwich ordering does not improve translation models, we show that they are robust to layer order changes, and that even extreme reorderings (all attention sublayers at the bottom, and all the feedforward sublayers at the top) perform as well as the baseline.

Sublayer reordering can improve the performance of transformer models, but an ordering that improves models on one group of tasks (word/character-level language modeling) might not improve the performance on another task. By showing that sublayer ordering can improve models at no extra cost, we hope that future research continues this line of work by looking into optimal sublayer ordering for other tasks, such as translation, question answering, and classification.

%% file: main.bbl
\begin{thebibliography}{28}
\expandafter\ifx\csname natexlab\endcsname\relax\def\natexlab#1{#1}\fi

\bibitem[{Ba et~al.(2016)Ba, Kiros, and Hinton}]{layernorm}
Jimmy~Lei Ba, Jamie~Ryan Kiros, and Geoffrey~E. Hinton. 2016.
\newblock Layer normalization.
\newblock {arXiv:1607.06450}.

\bibitem[{Baevski and Auli(2019)}]{baevski2018adaptive}
Alexei Baevski and Michael Auli. 2019.
\newblock \href {https://openreview.net/forum?id=ByxZX20qFQ} {Adaptive input
  representations for neural language modeling}.
\newblock In \emph{ICLR}.

\bibitem[{Correia et~al.(2019)Correia, Niculae, and
  Martins}]{correia2019adaptively}
Gon\c{c}alo~M. Correia, Vlad Niculae, and Andr\'{e} F.~T. Martins. 2019.
\newblock Adaptively sparse transformers.
\newblock {arXiv:1909.00015}.

\bibitem[{Dai et~al.(2019)Dai, Yang, Yang, Carbonell, Le, and
  Salakhutdinov}]{transformerXL}
Zihang Dai, Zhilin Yang, Yiming Yang, Jaime Carbonell, Quoc Le, and Ruslan
  Salakhutdinov. 2019.
\newblock \href {https://www.aclweb.org/anthology/P19-1285} {Transformer-{XL}:
  Attentive language models beyond a fixed-length context}.
\newblock In \emph{ACL}.

\bibitem[{Devlin et~al.(2019)Devlin, Chang, Lee, and Toutanova}]{BERT}
Jacob Devlin, Ming-Wei Chang, Kenton Lee, and Kristina Toutanova. 2019.
\newblock \href {https://doi.org/10.18653/v1/N19-1423} {{BERT}: Pre-training of
  deep bidirectional transformers for language understanding}.
\newblock In \emph{NAACL}.

\bibitem[{Guo et~al.(2019)Guo, Qiu, Liu, Shao, Xue, and
  Zhang}]{startransformer}
Qipeng Guo, Xipeng Qiu, Pengfei Liu, Yunfan Shao, Xiangyang Xue, and Zheng
  Zhang. 2019.
\newblock \href {https://doi.org/10.18653/v1/N19-1133} {Star-transformer}.
\newblock In \emph{NAACL}.

\bibitem[{Inan et~al.(2017)Inan, Khosravi, and Socher}]{inan2017}
Hakan Inan, Khashayar Khosravi, and Richard Socher. 2017.
\newblock \href {https://openreview.net/forum?id=r1aPbsFle} {Tying word vectors
  and word classifiers: {A} loss framework for language modeling}.
\newblock In \emph{ICLR}.

\bibitem[{Jozefowicz et~al.(2015)Jozefowicz, Zaremba, and
  Sutskever}]{jozefowicz2015empirical}
Rafal Jozefowicz, Wojciech Zaremba, and Ilya Sutskever. 2015.
\newblock An empirical exploration of recurrent network architectures.
\newblock In \emph{ICLR}.

\bibitem[{Khandelwal et~al.(2019)Khandelwal, Levy, Jurafsky, Zettlemoyer, and
  Lewis}]{urvashi}
Urvashi Khandelwal, Omer Levy, Dan Jurafsky, Luke Zettlemoyer, and Mike Lewis.
  2019.
\newblock Generalization through memorization: Nearest neighbor language
  models.
\newblock {arXiv:1911.00172}.

\bibitem[{Kuhn(1955)}]{kuhn1955hungarian}
Harold~W. Kuhn. 1955.
\newblock The hungarian method for the assignment problem.
\newblock \emph{Naval Research Logistics Quarterly}, 2(1-2):83--97.

\bibitem[{Lu et~al.(2019)Lu, Li, He, Sun, Dong, Qin, Wang, and
  Liu}]{lu2019understanding}
Yiping Lu, Zhuohan Li, Di~He, Zhiqing Sun, Bin Dong, Tao Qin, Liwei Wang, and
  Tie-Yan Liu. 2019.
\newblock Understanding and improving transformer from a multi-particle dynamic
  system point of view.
\newblock {arXiv:1906.02762}.

\bibitem[{Merity et~al.(2016)Merity, Xiong, Bradbury, and
  Socher}]{merity2016pointer}
Stephen Merity, Caiming Xiong, James Bradbury, and Richard Socher. 2016.
\newblock Pointer sentinel mixture models.
\newblock {arXiv:1609.07843}.

\bibitem[{Nguyen and Salazar(2019)}]{nguyen2019transformers}
Toan~Q. Nguyen and Julian Salazar. 2019.
\newblock Transformers without tears: Improving the normalization of
  self-attention.
\newblock {arXiv:1910.05895}.

\bibitem[{Ott et~al.(2018)Ott, Edunov, Grangier, and Auli}]{scalingnmt}
Myle Ott, Sergey Edunov, David Grangier, and Michael Auli. 2018.
\newblock \href {https://doi.org/10.18653/v1/w18-6301} {Scaling neural machine
  translation}.
\newblock In \emph{CMT}.

\bibitem[{Press and Wolf(2017)}]{press2017}
Ofir Press and Lior Wolf. 2017.
\newblock \href {https://www.aclweb.org/anthology/E17-2025} {Using the output
  embedding to improve language models}.
\newblock In \emph{EACL}.

\bibitem[{Radford et~al.(2018)Radford, Narasimhan, Salimans, and
  Sutskever}]{gpt-1}
Alec Radford, Karthik Narasimhan, Time Salimans, and Ilya Sutskever. 2018.
\newblock Improving language understanding with unsupervised learning.

\bibitem[{Radford et~al.(2019)Radford, Wu, Child, Luan, Amodei, and
  Sutskever}]{gpt-2}
Alec Radford, Jeff Wu, Rewon Child, David Luan, Dario Amodei, and Ilya
  Sutskever. 2019.
\newblock Language models are unsupervised multitask learners.

\bibitem[{Rae et~al.(2020)Rae, Potapenko, Jayakumar, Hillier, and
  Lillicrap}]{compressive}
Jack~W. Rae, Anna Potapenko, Siddhant~M. Jayakumar, Chloe Hillier, and
  Timothy~P. Lillicrap. 2020.
\newblock \href {https://openreview.net/forum?id=SylKikSYDH} {Compressive
  transformers for long-range sequence modelling}.
\newblock In \emph{ICLR}.

\bibitem[{Sanh(2019)}]{sanh_2019}
Victor Sanh. 2019.
\newblock \href {https://medium.com/huggingface/distilbert-8cf3380435b5}
  {Smaller, faster, cheaper, lighter: Introducing {DistilBERT}, a distilled
  version of {BERT}}.

\bibitem[{Sennrich et~al.(2016)Sennrich, Haddow, and Birch}]{bpe}
Rico Sennrich, Barry Haddow, and Alexandra Birch. 2016.
\newblock \href {https://doi.org/10.18653/v1/p16-1162} {Neural machine
  translation of rare words with subword units}.
\newblock In \emph{ACL}.

\bibitem[{So et~al.(2019)So, Le, and Liang}]{evolvedtransformer}
David So, Quoc Le, and Chen Liang. 2019.
\newblock \href {http://proceedings.mlr.press/v97/so19a.html} {The evolved
  transformer}.
\newblock In \emph{ICML}.

\bibitem[{Srivastava et~al.(2014)Srivastava, Hinton, Krizhevsky, Sutskever, and
  Salakhutdinov}]{dropout}
Nitish Srivastava, Geoffrey Hinton, Alex Krizhevsky, Ilya Sutskever, and Ruslan
  Salakhutdinov. 2014.
\newblock \href {http://jmlr.org/papers/v15/srivastava14a.html} {Dropout: A
  simple way to prevent neural networks from overfitting}.
\newblock \emph{Journal of Machine Learning Research}, 15:1929--1958.

\bibitem[{Sukhbaatar et~al.(2019)Sukhbaatar, Grave, Bojanowski, and
  Joulin}]{Sukhbaatar2019}
Sainbayar Sukhbaatar, Edouard Grave, Piotr Bojanowski, and Armand Joulin. 2019.
\newblock \href {https://doi.org/10.18653/v1/p19-1032} {Adaptive attention span
  in transformers}.
\newblock In \emph{ACL}.

\bibitem[{Tan and Le(2019)}]{efficientnet}
Mingxing Tan and Quoc Le. 2019.
\newblock \href {http://proceedings.mlr.press/v97/tan19a.html}
  {{E}fficient{N}et: Rethinking model scaling for convolutional neural
  networks}.
\newblock In \emph{ICML}.

\bibitem[{Vaswani et~al.(2017)Vaswani, Shazeer, Parmar, Uszkoreit, Jones,
  Gomez, Kaiser, and Polosukhin}]{AIAYN}
Ashish Vaswani, Noam Shazeer, Niki Parmar, Jakob Uszkoreit, Llion Jones,
  Aidan~N Gomez, \L~ukasz Kaiser, and Illia Polosukhin. 2017.
\newblock \href
  {http://papers.nips.cc/paper/7181-attention-is-all-you-need.pdf} {Attention
  is all you need}.
\newblock In \emph{NeurIPS}.

\bibitem[{Zhang et~al.(2019)Zhang, Titov, and Sennrich}]{zhang2019improving}
Biao Zhang, Ivan Titov, and Rico Sennrich. 2019.
\newblock Improving deep transformer with depth-scaled initialization and
  merged attention.
\newblock {arXiv:1908.11365}.

\bibitem[{Zhu et~al.(2015)Zhu, Kiros, Zemel, Salakhutdinov, Urtasun, Torralba,
  and Fidler}]{moviebook}
Yukun Zhu, Ryan Kiros, Richard Zemel, Ruslan Salakhutdinov, Raquel Urtasun,
  Antonio Torralba, and Sanja Fidler. 2015.
\newblock Aligning books and movies: Towards story-like visual explanations by
  watching movies and reading books.
\newblock {arXiv:1506.06724}.

\bibitem[{Zoph and Le(2016)}]{zoph2016neural}
Barret Zoph and Quoc~V. Le. 2016.
\newblock Neural architecture search with reinforcement learning.
\newblock {arXiv:1611.01578}.

\end{thebibliography}
